# Small Object Detection using Deep Learning

Aleena Ajaz, Ayesha Salar, Tauseef Jamal, Asif Ullah Khan
DCIS, PIEAS University, Pakistan

*Abstract*— **Now-a-days, UAVs such as drones are greatly used for various purposes like that of capturing and target detection from ariel imagery etc. Easy access of these small ariel vehicles to public can cause serious security threats. For instance, critical places may be monitored by spies blended in public using drones. Study in hand proposes an improved and efficient Deep Learning based autonomous system which can detect and track very small drones with great precision. The proposed system consists of a custom deep learning model 'Tiny YOLOv3', one of the flavors of very fast object detection model 'You Look Only Once' (YOLO) is built and used for detection. The object detection algorithm will efficiently the detect the drones. The proposed architecture has shown significantly better performance as compared to the previous YOLO version. The improvement is observed in the terms of resource usage and time complexity. The performance is measured using the metrics of recall and precision that are 93% and 91% respectively.**

## I. Introduction

As the world advances while incorporating newest technologies, security systems are also evolving at a rapid pace to utilize the power of innovation to tackle any security threats that may arise.

The use of drones has been increasing tremendously for the past few years including accessibility to public which has further increased its usage. With the increase in usage of drones, the increase in security threats has also increased. We have seen some incidents in which the drones are flying over different buildings including some of the sensitive areas which require security. Drones also propose many threats including drones being used as a drug smuggling tool, hacking the drones causing them to either explode or using them to steal sensitive information, drones are also being used to do illegal surveillance causing safety and privacy threats. These events have raised great threat to safety and privacy. So, the solution to these security issues is to detect and track these drones so counter techniques can be done to stop these drones to ensure safety and privacy.

Moreover, majority security or defense systems have the capability to detect the magnitude of threat that may occur however, very few of them have the capability to analyze the threat in depth. As Machine Learning and Artificial Intelligence advance to make such systems intelligent by reducing the human intervention, effective object detection can help make the security systems more efficient and perform better analysis.

There are different techniques to detect and track drones and many systems using these techniques can be found in the market. Almost all of these systems provide the consumer with the ability to autonomously detect and track the drones. These systems include many different technologies such as RF (Radio Frequency) signal detection in which the system uses electromagnetic waves for the communication between the ground operators and the device. These RF signals carry the information from one point to another through air. Some systems use LIDAR (Light detection and ranging) in which light is used in the form of pulsed laser for the measurement of different ranges. LIDAR provides a remote sensing through which the communication ground operator and the drone. While some systems use RIDAR in which the system detects the drone by using Radio Frequency waves. It emits the short pulses of RF signals. And it performs the detection by the reflected signals it receives from the drone. Other system use acoustics which uses multiple microphones and cameras to monitor the drones. The system then uses these images and audios captured for the monitoring of drones.

Now coming to the reason why these approaches are not feasible. The main issue in LIDAR is it only detects the drones when the drone's movements and the LIDAR sensor's capabilities corresponds with each other. And the fact it is a new technology its accuracy and feasibility is still a question. It's not cost effective which is another reason why it is unfeasible.

The issue with RIDAR approach is that it can gives us false positives and it is not feasible for detecting the small drones. As these drones are flying at relatively lower velocities, it decreases the Doppler signature. Thus, it is not feasible for the detection of small drones especially for autonomous detection.

With RF signal detection, the main issue that arises is that the drones can be operated without the ground operator, but we have to use pre-programmed flight path. In systems using acoustic detection they do not provide us with higher accuracy and a feasible range. The maximum range of detection for these systems is below 200–250 thus making it unfeasible. These systems also cannot operate in noisy areas.

Nowadays, the technique of object detection integrated in a security system has emerged as potential way to detect objects that may pose threats. One of the most feasible approaches to detect and track the drones is using Optics. It is the most convenient of all the methods we discussed above. In optics cameras are used to provide input in the form of images. In this approach we have many advantages which we did not have in the above discussed approaches. It provides us with robustness, convenience, accurate results and a large range. Using this approach along with the deep learning algorithms provides us with the most convenience and helps us tackle the issues we encountered in other approaches.

There are many deep learning algorithms that can be used to perform detection. Object detectors are used for detection. There are many object detectors like Res-Net, CNN, YOLO etc. YOLO is a real time object detector which is also called a single shot detector meaning that it detects objects in one single shot. One of these detectors is YOLOv3 which is a type of YOLO object detector. It uses multi scale prediction, Darknet 53 as its backbone classifier, a strong feature extractor and it provides much better accuracy than its previous versions. There are many versions of YOLOv3 one of which is Tiny YOLOv3 which includes a small light weight architecture which is much faster than YOLOv3. The main drawback of Tiny YOLOv3 is that it does not perform better accuracy than YOLOv3 as it performs detection at 2 scale unlike YOLOv3 which performs detection at 3 scale.

The increasing use of drones by the public is a major threat to general security and privacy. In recent years, security incidents involving drones have been discovered such as, multiple drones wandering in the vicinity of nuclear power plants in France, terrible crash of drone near LAX airport, alarming events near White House, smuggling of drugs using drones and people caught spying on the swimming pools using drones etc. Despite their mobility and small size there should be a drone detection system installed in security places. Several systems and architectures proposed by in recent years provide autonomous drone detection and identification, which is a very important operational function. However, with conventional methods, it is impossible to detect small commercial drones.

Among other approaches, deep learning-based detection is the most effective and accurate [1]. We intend to propose a more robust and effective method for detection of drone by integrating artificial intelligence. This system which will greatly benefit the security systems of the country and prevent the attackers from exploiting any possible security vulnerabilities. Hence in our work, high performance of the deep learning algorithm for object detection for this very specific task is achieved.

## II. LItRATURE REVIEW

Computer vision undergoes typical problems when it comes to object detection like what is the object and where it is. Recognizing the object along with the position of object is more complex than classification problem. Also, when an image contains more than one image then classification does not work there.

### A. CNN

In neural networks, different task such as image classification, image identification, object detection is performed using Convolutional Neural Networks (CNNs). CNNs are exceptionally capable for processing data in the form of grids in which its structure holds the information about data, such as audio signals data in 1D, data in images in 2D, and videos in 3D. For classifying images, CNN takes an input image, process it, and classify it under certain categories (E.g., Airplane, Car, Truck etc.). Basically, if we give an image to the machine as an input then machines take it as an array of picture elements which are basically pixels, and these pixels vary with the resolution of image based on which machine sees its Height*Width* Dimension.

CNNs comprises of piles of layers which includes convolutional layers and pooling layers which incrementally learn multifaceted-level features in the form of filters. So, technically, deep learning CNN models are trained in a way that each image is moved across the series of different CNN layers with kernels, pooling layers, fully connected layers (FC) and at the end object is classifies by using SoftMax function is applied which decides the category of the object on the b-asis of probabilistic values which are between 0 and 1.

Each layer of CNN architecture and their functionality is discussed in the below paragraphs.

### 1) Convolution Layer

Convolution layer is the initial layer which receives an input image and extracts the features from that image [2]. It uses small squares of input data and learns the image feature by sustaining the correlation between image pixels' values. This layer basically is given two piece of data inputs which inlcude image matrix and filter matrix and perform mathematical operation to learn the features.

An Image matrix having dimension (h x w x d) where h= Height of the image, w= Width of the image, d= Dimension of the image.

A filter matrix of dimension (fh, x fw x d) where fh = Filter height, fw = Filter width, d=Dimension

After performing mathematical sliding operation, the output of the image matrix will become as (h – fh +1) x (w – fw + 1) x1.

In convolution, basically the filter matrix slides by pixel shift over the image and multiplies it with the image matrix which is technically known as convolution of an image. Convolution of m x n image matrix multiplies with p x q filter matrix gives us an output matrix called "Feature Map". Filters perform convolution on image in several ways for different purposes for example blurring, sharpening, edge prominence and detection and it all depends on the adjustment filter matrix.

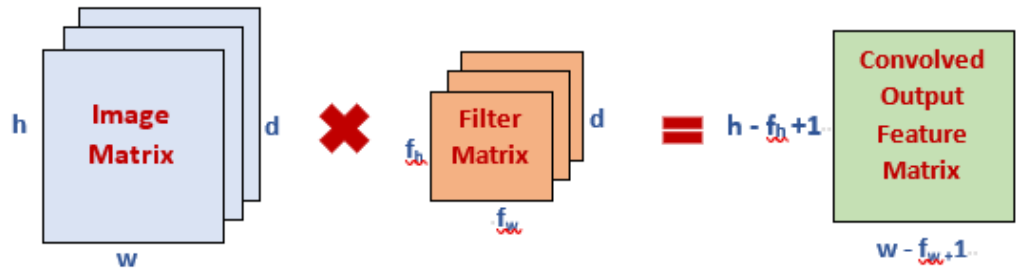

**Figure 1: Image matrix convolution operation with multiplies filter matrix.**

*2) Stride*

Stride works in specifying the numeral pixels shifts to be done when filter matrix slides onto the input image matrix. When stride is used as 1 then the filter matrix slides with 1 pixel move over the image at one time. When stride is used as 2 then the filter matrix slides with 2 pixel move over the image at one time

Let us see how the convolution of an image matrix of 6 x 6 with 3 x 3 filter matrix using stride 2 works.

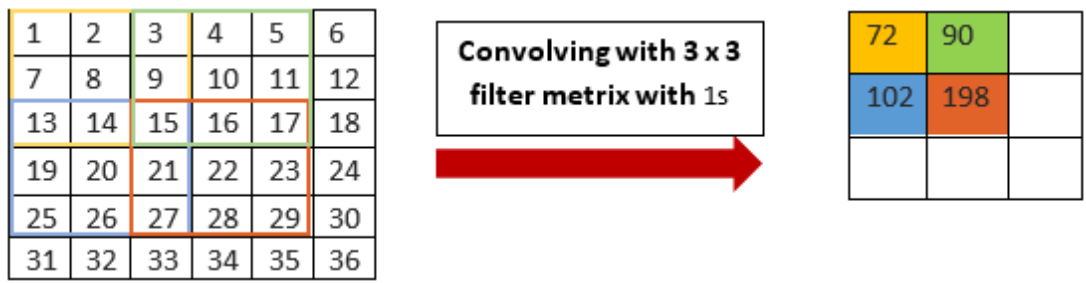

**Figure 2: Convolution with 3 x3 filter using Stride 2**

*3) Padding*

Padding is used when the filters do not entirely fit the image. So, we can do padding in two ways:

Zero-padding: Adding zeros to the picture to fit the filters.
Valid Padding: Discard the part of image which is not fit by the filter. This type of padding only holds the useable part of the image.

*4) Non-Linearity (ReLU)*

ReLU stands for Rectified Linear Unit. It is one of the functions that is used to add non-linearity. Its formula is:
$R(z) = \max(0,z)$

ReLU function is mostly used to bring in non-Linearity in the learning of CNNs. Because, in practical scenario, the data always wants the CNN to learn non-negative linear values.

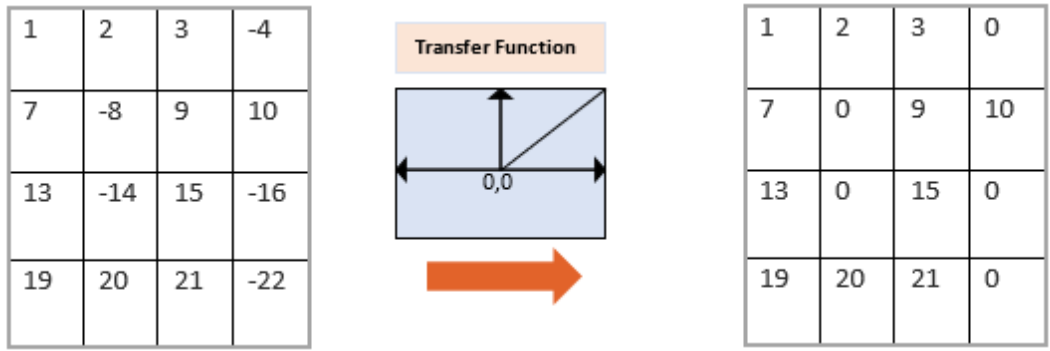

Figure 3 ReLU operation

ReLU is proved to be a quicker activation function in terms of learning [4] and it is also proved as to most effective and broadly utilized function [5]. It offers better performance and generalization in DL compared to other AFs [6].

*5) Pooling Layer*

Sometime, the images are huge therefore we reduce some parameters of image by using pooling layers. Pooling layers does not discard the important parameters. Spatial pooling, also known as subsampling or down sampling has various types including:
- Max Pooling
- Average Pooling
- Sum Pooling

In Max pooling, we include the highest element from the rectified feature map.
In Average pooling, we include average of elements from the rectified feature map.
In Sum pooling, we keep the summation of all elements in the feature map.

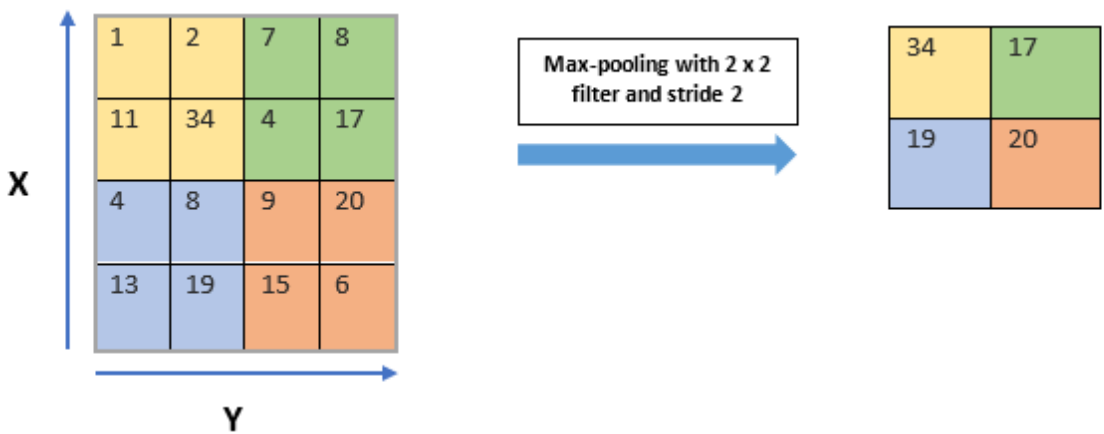

Figure 4: Max-Pooling

*6) Fully Connected Layer*

The last few layers are constituted by Fully Connected Layers. Basically, the output feature maps from final pooling or CNN layer are flattened which means it will be converted to vector and then FC layers takes that flattened output as input.

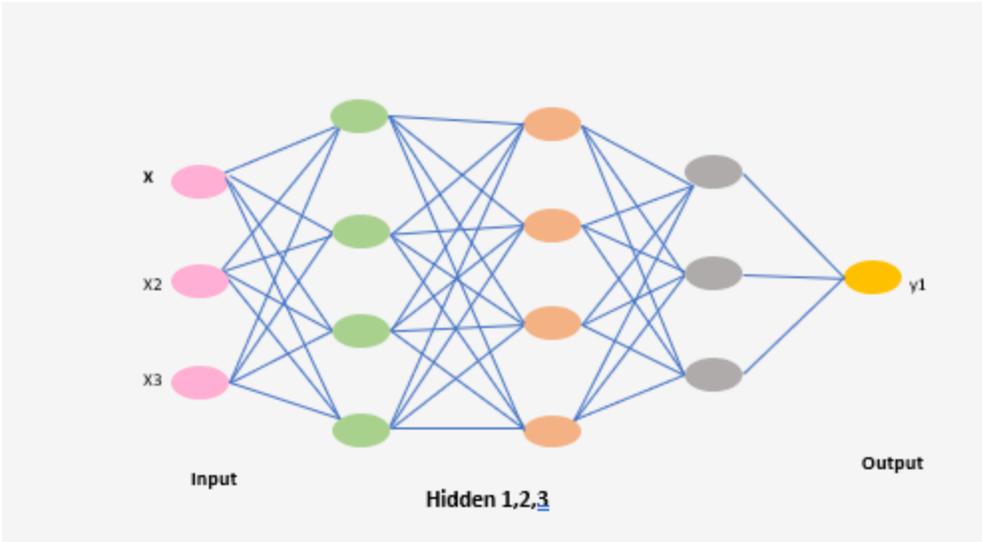

**Figure 5:Fully Connected Layer**

In the end, we apply activation function i.e., SoftMax or sigmoid to classify the objects to determine their category.

*B. YOLO*

There are different algorithms for Object detection problems. One of them is YOLO which stands for YOU ONLY LOOK ONCE.

YOLO is a smart convolution neural network (CNN) for detecting objects in real-time[6]. In YOLO, only a single network is applied to the whole image, and then image is divided into SXS grid cells. For each cell, certain number of bounding boxes are to be predicted and after that, class probabilities for each box are predicted which decides the class of the objects in that bounding box[7]. YOLO actually "looks once" to the image in a way that it entails the neural network to pass one forward propagation to make predictions. It means that prediction of several bounding boxes and probabilities of their classes happens simultaneously[8]. That is why, it is also known as "a single shot detector" After making sure that it has detected each object once, it will result into detected objects along with the bounding boxes.

*1) How does YOLO framework works?*

Let us see the brief functionality of YOLO framework. Following steps mentions the working of YOLO for detecting the objects in each image.

1. First YOLO takes input as an image.
2. Then, the image is divided into RXR grid for example 3 x 3 grid.

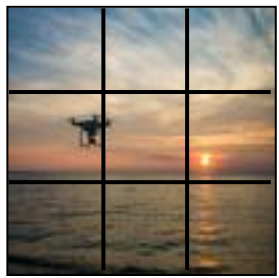

3. On each grid, classification and localization will be applied and then predictions of bounding boxes with class probabilities of respective bounding boxes for objects will be given by YOLO.

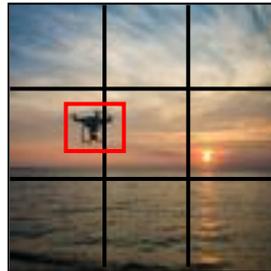

Now, we understand how this model is trained. To train YOLO, labelled data is passed to the model. Let us consider that image is divided into a grid of 3 x 3 size and there is one class of object to be classified. Our class is "Drone". Therefore, there will be a 'y' label for each cell of grid and in above case the 'y' label will be six-dimensional vector.

$$y = \begin{bmatrix} pc \\ bx \\ by \\ bh \\ bw \\ c \end{bmatrix}$$

Above table shows,

- $p_c$ identifies the probability of presence of object in the grid cell.
- $b_x, b_y, b_h, b_w$ shows parameters of bounding box including its x-coordinate, y-coordinate, height, and width if there is an object.
- 'c' is the class of the object if there is any. So, in our case, the object is a 'Drone', which means c will 1.

If we see our first grid from above example which is: suppose we choose the first grid cell from above:

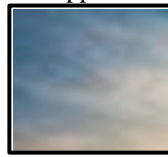

As seen, there is no object present in above grid cell, so 'pc' will be 0 and the 'y' label for this grid cell will be:

$$y = \begin{bmatrix} 0 \\ ? \\ ? \\ ? \\ ? \\ ? \end{bmatrix}$$

'?' implies that whatever the value of $b_x$, $b_y$, $b_h$, $b_w$, c is, it does not matter because there is no object in that grid cell.

Now, if we see another grid cell in which we have our object 'Drone' (c = 1):

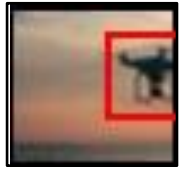

Before writing the y label for above cell, yolo will first understand if there is object present or not in the cell. Yolo will take the midpoint of the objects and then they will be allotted to the cell that contains the midpoint of that object. So, the label-y for this cell having drone would be:

$$y = \begin{bmatrix} 1 \\ bx \\ by \\ bh \\ bw \\ 1 \end{bmatrix}$$

Probability of objectness '$p_c$' will be 1 as an object is present in this cell. The attributes of the bounding box $b_x$, $b_y$, $b_h$, $b_w$ will be computed with respect to the exact cell we are looking at. Since the class is only one which is 'drone', so class probability will be c=1. So, for each of the 9 cells, we will have a 6-dimensional output vector. This output will have a shape of 3 X 3 X 6.

Now, we have an input image and its target vector. In case of our above example, we have input as 100x100x3 and corresponding output vector as 3x3x6.

*2) How Bounding Boxes are encoded?*

Attributes of bounding boxes including x coordinate, y coordinates, height, and width bx, by, bh, and bw are calculated with respect to the grid cell we are looking at. For better understanding, consider the grid cell which contains our object which is drone:

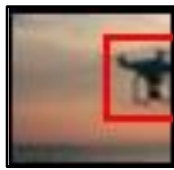

So, $b_x$, $b_y$, $b_h$, and $b_w$ will be computed with respect to this grid only. The 'y' label for this grid will be:

$$y = \begin{bmatrix} 1 \\ bx \\ by \\ bh \end{bmatrix}$$

|   | bw |
|---|----|
|   | 1  |

Objectness probability is pc = 1 as an object is present in this cell and as the object is a drone, so c= 1. Here the question is that how the bounding box values are decided? Let us see how these values are encoded bx, by, bh, and bw .In YOLO, all the cells are given with the coordinates as shown below:

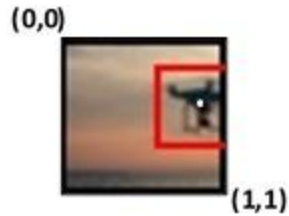

Basically bx, by are the x, y coordinates of the object's midpoint with respect to the cell. So these values will be approximately around bx = 0.8 and by = 0.4:

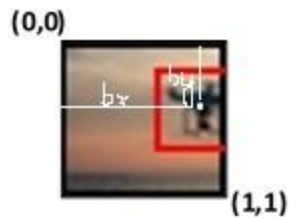

Now coming towards height, bh is the fraction of the bounding box's height to the height of the equaling cell, which in our case is around 0.8. So, bh = 0.8. and similarly, bw is the fraction of the bounding box's width to the width of the matching cell. Here, bw = 0.5 (approximately). The 'y' label will be:

| y= | 1 |
|----|---|
|    | 0.8 |
|    | 0.4 |
|    | 0.8 |
|    | 0.5 |
|    | 1 |

It is important to understand that attributes of bounding boxes bx and by will always be between 0 and 1 as the midpoint will always lie inside the cell.

*3) Intersection over Union and Non-Max Suppression*

It is a fraction of the overlap area of predicted bounding box with the area of actual bounding box. Let us see the actual and predicted bounding boxes for a drone as:

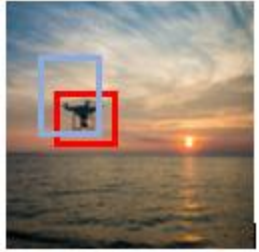

Above, the red box demonstrates the actual bounding box and the blue box is the predicted one. So, how it is decided that the prediction of box is good or not? Basically, area of the intersection over union of these two boxes will be determined by IOU which will be:

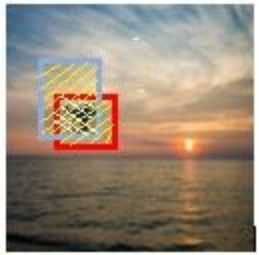

IoU = Area of white filled box / Area of yellow filled box

Basically, we set a threshold that if If IoU is greater than 0.5, then it is accepted that the prediction is good enough. This value can be increased or decrease based on the problem. Instinctively, the more threshold is increased, the better the predictions become.

*4) Techniques to improving YOLO performance*

Here is another technique which is used to expand the performance of yolo. It is Non-Max Suppression through which it is decided which predicted box should be decided as result[9].
- **Non-max Suppression**

Most of the time, object detection algorithms encounter a problem that besides identifying an object one time, it might perform detection several times.

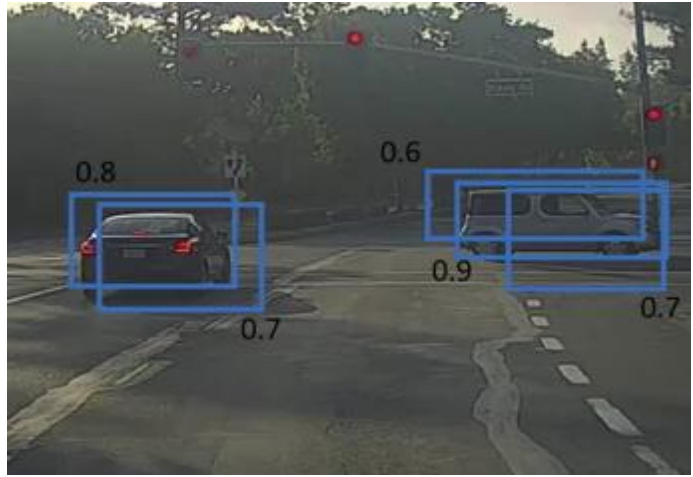

In the above detections the drones are recognized more than one time. The Non-Max Suppression technique polishes up this to get a single detection per object. So, we will be keeping the boxes with highest probability and discard the boxes with non-max probabilities.

- **Anchor Boxes**

Each grid can only identify one object but there can be a case if there are more than one object in a single grid cell. In that case, the concept of anchor boxes comes. Let us say that the below image is divided into a 3 X 3 grid:

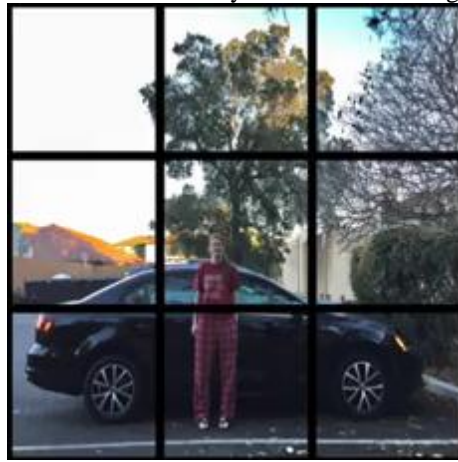

To assign a cell with an object, the midpoint of the object will be taken and based on that midpoint, a grid cell will be assigned to that object. In above case, both objects have midpoint lie in the same cell. So, definite bounding boxes for the objects will be like:

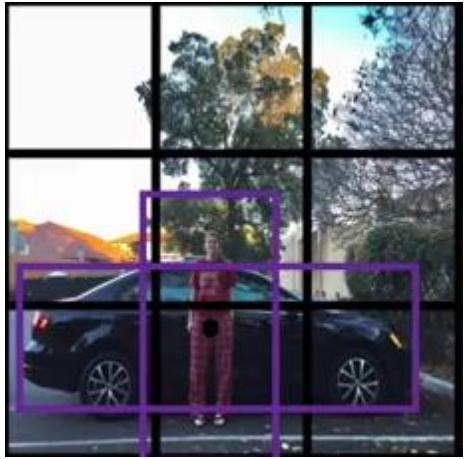

From the above logic of non-max suppression procedure, we will get only one box per grid cell, either for the person or for the car. In this case, we must use anchor boxes which will be able to output both boxes from one grid cell.

5) *How anchor boxes work?*

First, n number of shapes will be predefined which are called anchor box shapes. And, for each cell, as a replacement for having one output, we have n detections. In the above case, we will predefine two anchor boxes.

Anchor box 1:    Anchor box 2:

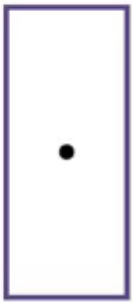 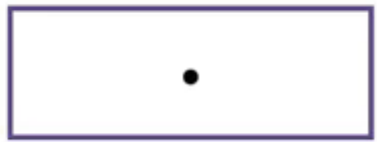

So, 'y' label for YOLO excluding anchor boxes will be like:

$$Y = \begin{bmatrix} pc \\ bx \\ by \\ bh \\ bw \\ c1 \\ c2 \end{bmatrix}$$

c1 and c2 indicated 2 classes (person and car) for above case.

Now, if we have predefine 2 anchor boxes then our y label will be:

$$y = \begin{bmatrix} p_c \\ b_x \\ b_y \\ b_h \\ b_w \\ c_1 \\ c_2 \\ p_c \\ b_x \\ b_y \\ b_h \\ b_w \\ c_1 \\ c_2 \end{bmatrix}$$

The first 7 rows of y label correspond to anchor box 1 and the remaining 7 correspond to anchor box 2. Anchor boxes will be assigned object based on their similarity of their shape and bounding boxes. As the shape of anchor box 1 is like the bounding box for the person, so the object person will be assigned to the anchor box 1 and the other object which is a car will be given to anchor box 2. In this case our output vector will be changed. The output vector will be 3 X 3 X 14 (as we have 2 anchors and 3 X 3 grid with 2 classes).
With the concept of anchors, we can detect more than one objects in one grid cell based on the anchors numbers we define.

*6) Algorithm steps*

Simple algorithm steps of how YOLO works:

- Receives an image of shape (416, 416, 3) as input.
- Give this image to a CNN, which returns a (13, 13, 5, 6) dimensional output.
- From the above, the final two dimensions are leveled to get an output of (13, 13, 30):
    o Here, each cell of a 13 X 13 grid returns 30 numbers.
    o 30 = 5 * 6, where 5 indicates the number of anchor boxes per grid cell.
    o 6 = 5 + 1, 5 representing the attributes pc, bx, by, bh and bw. The number of classes for detection is represented by 1.
- Finally, IoU and Non-Max Suppression is performed to avoid selecting overlapping boxes.

*C. YOLOv3*

This version of YOLO is also a single shot detector but with a higher speed, a better performance and an improved training. It uses multi scale prediction, Darknet 53 as its backbone classifier, a strong feature extractor and it provides much better accuracy than its previous versions.

The previous versions of YOLO had difficulty detecting small objects but YOLOv3 provides a better small object detection. The main purpose of Darknet-53 is to perform feature extraction. Initially it contained 53 convolutional layers which are trained on ImageNet. These convolutional layers are of stride 2 so that the feature maps are down sampled. These convolutional layers are not followed by max pooling layers as they exclude low level feature maps. Instead, the max pool layers, these convolutional layers are followed by batch normalization which ensures the inputs are normalized even in deeper layers and Leaky RELU which help in avoiding over-fitting. In addition to that it has 53 more layers which are stacked on the previous layers for the purpose of detection. YOLOv3 captures low level features through the additional convolutional layers which helped in improving the small object detection. Thus, the YOLOv3 has a total of 106 layers. This is one of the reasons why it has a lower speed due to its complex underlying architecture.

YOLOv3 performs detection at layers 82, 94 and 106 performs detection at three different scales. This is achieved through the network down sampling the input image by factors of 32, 16 and 8. These factors are known as the stride of the network which show us how the input at the three layers 82, 94, 106 is smaller than the input of the network. Let us take an example if the input size of the network is 416 x 416 and the stride is 32 then the output will be of size 13x3. Subsequently for the stride 16 output will be of size 26 x 26, for stride 8 the output will be of size 52x52. In the architecture of YOLOv3, 13x13 detects the larger objects, 52 x 52 detects the smaller objects and 26x26 detects medium objects.

On the feature maps it applies a 1x1 detection kernels where three different sizes are used at three different places in the network i.e. 1x1 kernel applied to down-sampled images to the outputs of sizes discussed above. The detection kernel shape has a depth which is calculated by: 1 x 1 x (B x (5 + C)) where B represents the total amount of bounding boxes, C represents the number of classes on which we are training our network. For each cell on the feature map, YOLOv3 predicts a total of 3 bounding boxes. For each bounding box the attribute 5 + c describes the bounding box's center coordinates and dimensions (width and height), object score and a list where all confidences for each class this bounding box might belong to.

As mentioned earlier YOLOv3 for each cell of the feature map, it predicts 3 bounding boxes. In return an object is predicted by every only if the center of that object belongs to the cell. YOLOv3 detects the cell which has the center of the object. While YOLOv3 is being trained, it has a total of 1 ground truth bounding box which is used to detect one object. To define to which cell this bounding box belongs to is the most important step.

If the detection scale let's say is 32 which is also the stride and the input of the image is 416*416 then the image will be down sampled by 13 by 13 grid cells. This will show us the output feature map. The ground truth to which these cells belong to are identified and the network assigns the responsibility to predict the object to the center of the cell. The object score for this particular cell will be 1. All the cells during the training predict 3 bounding boxes

For the purpose of predicting bounding boxes, YOLOv3 uses predefined default bounding boxes called anchor boxes. The purpose of these anchor boxes is to predict bounding boxes' real width and height. It uses 9 anchor boxes where for each scale 3 anchor boxes are used. Thus for every scale each cell of the feature map will predict 3 bounding boxes utilizing the 3 anchors.

YOLOv3 predicts a massive amount of bounding boxes unlike the previous versions. That means YOLOv3 will predict boxes at different scales which is also one of the reasons why it is slower than the previous versions. Now how to choose the bounding box that best predicts the object in question. In YOLOv3 the bounding box having an anchor which has the highest IoU with the ground box is responsible for detecting the object.

It performs multi label classification and does not use soft maxing for the class predictions like in the previous versions. Soft maxing was the rule that the classes are mutually exclusive which is not true for datasets like Person or Women. Instead, it predicts the object confidence and the class predictions through logistic regression where the threshold is used for predicting multi labels for an object which means that the classes of the objects with a greater score than the threshold are allocated to the box [10] [11].

*D. Tiny YOLOv3*

There is a less complex version of YOLOv3 which is called Tiny YOLOv3. As discussed previously YOLOv3 has 106 layers, in comparison to it the tiny YOLOv3 architecture has only

24 layers including 7 convolutional layers and 6 max pooling layers. Tiny YOLOv3 is also a real time object detector like YOLOv3 and due to its lighter architecture is suitable for devices with poor data processing properties. It is much faster than YOLOv3 due to its simplified architecture and the major advantage is that it does not occupy large amount of memory, but the tradeoff is that it does not have a better accuracy. This architecture does not detect small objects due to the property of Tiny YOLOv3 performing detection at two scales i.e. at 13*13 and 26*26 scale. Unlike YOLOv3, Tiny YOLOv3 defines its loss function according to the bounding box position error, error of bounding box and error of classification prediction [12].

**Figure 6 Yolov3 Architecture**

| | Type | Filters | Size | Output |
|---|---|---|---|---|
| | Convolutional | 32 | 3 × 3 | 256 × 256 |
| | Convolutional | 64 | 3 × 3 / 2 | 128 × 128 |
| | Convolutional | 32 | 1 × 1 | |
| 1× | Convolutional | 64 | 3 × 3 | |
| | Residual | | | 128 × 128 |
| | Convolutional | 128 | 3 × 3 / 2 | 64 × 64 |
| | Convolutional | 64 | 1 × 1 | |
| 2× | Convolutional | 128 | 3 × 3 | |
| | Residual | | | 64 × 64 |
| | Convolutional | 256 | 3 × 3 / 2 | 32 × 32 |
| | Convolutional | 128 | 1 × 1 | |
| 8× | Convolutional | 256 | 3 × 3 | |
| | Residual | | | 32 × 32 |
| | Convolutional | 512 | 3 × 3 / 2 | 16 × 16 |
| | Convolutional | 256 | 1 × 1 | |
| 8× | Convolutional | 512 | 3 × 3 | |
| | Residual | | | 16 × 16 |
| | Convolutional | 1024 | 3 × 3 / 2 | 8 × 8 |
| | Convolutional | 512 | 1 × 1 | |
| 4× | Convolutional | 1024 | 3 × 3 | |
| | Residual | | | 8 × 8 |
| | Avgpool | | Global | |
| | Connected | | 1000 | |
| | Softmax | | | |

*E. Darknet*

Darknet is known as an open-source neural network framework which provides higher accuracy and fast rate for real-time object detection. For increasing accuracy of a custom object detection models, darknet was written in C and CUDA technology which allows the model to make real time predictions. It supports GPU computation [13].

Compute Unified Device Architecture (CUDA) is a platform with parallel computing and programming model which helps in accelerating the computational rate by utilizing GPU. It was created by Nvidia in 2006 and has been used in different applications to get the benefit of faster computation by using GPU[14].

It is used for object detection and provides us with various features which makes it faster and better than other neural network architectures. There are two optional dependencies when installing darknet i.e., OpenCV or CUDA. If there is a need for a wider range of different image types to be supported, then one can choose OpenCV and if there is a need to have GPU computation then one can choose CUDA[15].

In a feature called Nightmare, Darknet can also be used to run NN backwards. It can also handle recurrent neural networks, very powerful models in which the data being represented is changed over time, without using OpenCV or CUDA[15].

Darkent-53 is a CNN which contains a total of 53 layers. This is used as a backbone for YOLOv3 which is a real time object detector. The main purpose of Darknet-53 is to perform feature extraction. Compared to different backbones like ResNet 101 and ResNet 152, Darknet-53 provides better results and performance. We can use the pertained version of Darknet-53 which can classify images into a lot more object categories [16].

*1) Installing Darknet (https://pjreddie.com/darknet/install/):*

As mentioned above, there are two optional dependencies when installing darknet i.e. OpenCV or CUDA. If there is a need for a wider range of different image types to be supported, then one can choose OpenCV and if there is a need to have GPU computation then one can choose CUDA. These both options are optional.

To build Darknet, we first need to install the base system. To install the base system, we need to follow the steps given below:

The first step is to clone the Darknet repository which is available on Github by:

**git clone https://github.com/pjreddie/darknet.git**
**cd darknet**
**make**

The second step is to build darknet by: **./darknet**

*2) Utilizing CUDA*

As mentioned above, using CUDA provides us with the benefit to perform computations on GPU. Using GPU, we can accelerate our computational power. It stands for Graphic Processing Unit which is generally used to render pixels which provides better graphics in addition to a high frame rate.

To install CUDA, we can simply install it by specifying our operating system, architecture, version and installer type in this link: [17].

Now to enable CUDA we need to **make** the project. And if after installing CUDA, one wants to utilize CPU, run this command:

**./darknet -nogpu imagenet test cfg/alexnet.cfg alexnet.weights**

*3) Utilizing OpenCV*

Even though Darknet supports a lot of image formats but if one wants the framework to support more images formats like CMYK jpegs etc. then OpenCV will be used.

OpenCV can be installed from this link with the help of package manager: https://opencv.org/

In the Makefile, change the second line to OPENCV=1 and then remake the project.

III. PROPOSED SOLUTION

*A. Data Set*

Collecting data is perhaps the most important job for training a machine learning algorithm. As far as material is concerned, an open source labeled dataset was acquired for training our custom object detector. On this dataset, previously a system 'Drone-Net' was retrained [18]. This dataset consisted of 2664 images of DJI drones with normalized labels. Label file contained box coordinates which must be normalized. So, corresponding labels of images contained specifications as object-id, which is drone, normalized x_center, y_center, width and height of the box.

The bounding box annotations for the objects is also present in the dataset. In a label file, every row signifies 1 bounding box in the image.

One of the first steps for any machine learning training process is to split data into two sets randomly.

Train set: This set contains data which is used for training our model. For training our model we randomly selected between 70% to 90% of the data for training.
Test set: This set contains data which is used when for validating our model. But an important condition for data to be part of testing set, it shouldn't be part of the training set. Typically, this set is selected with 10-30% of the total data.

### B. Proposed Architecture

As discussed above even though Tiny YOLOv3 is much faster than YOLOv3, it does not give a better accuracy since it performs detection at 2 scale. To better achieve the accuracy, we are introducing a modified Tiny YOLOv3 architecture. This architecture includes a total of 31 layers: 16 convolutional layers, 3 detection(yolo) layers and rest are layers for max-pooling and routing. Previous YOLOV3 tiny architecture consisted of total 24 layers including 13 convolutional layers, 3 detection (yolo) layers. We have modified the Tiny YOLOv3 architecture in such a way that detections will be made on three scale by using the route layer from the previous layers to extract the features. It means that 3 yolo layers will be responsible for detecting the objects. Anchor boxes are used to calculate predicted bounding boxes width and height. Usually, YOLOv3 uses 3 anchors for each cell which predict three bounding boxes making a total of 9 anchor boxes. For our custom object detector, we have used a total of 6 anchor boxes[10]. Our prediction feature maps will be 13x13, 26x26 and 52x52 on 3 scales, respectively. As our network propagate the image ahead, at the first yolo layer, we get an output feature map of 13x13. After the placement of first yolo layer, we up sampled the next layers by a factor of two and then we concatenated with feature maps of previous layers having alike sizes. For second yolo layer, we get the output feature map of 26x26 and at third yolo layer we get a feature map of 52x52. This architecture is more concise and smaller than the original yolov3 architecture which do detections at 3 scale. Also, our small and light weighted architecture can detect small objects with high confidence score.

**Figure 7 Our Custom Architecture**

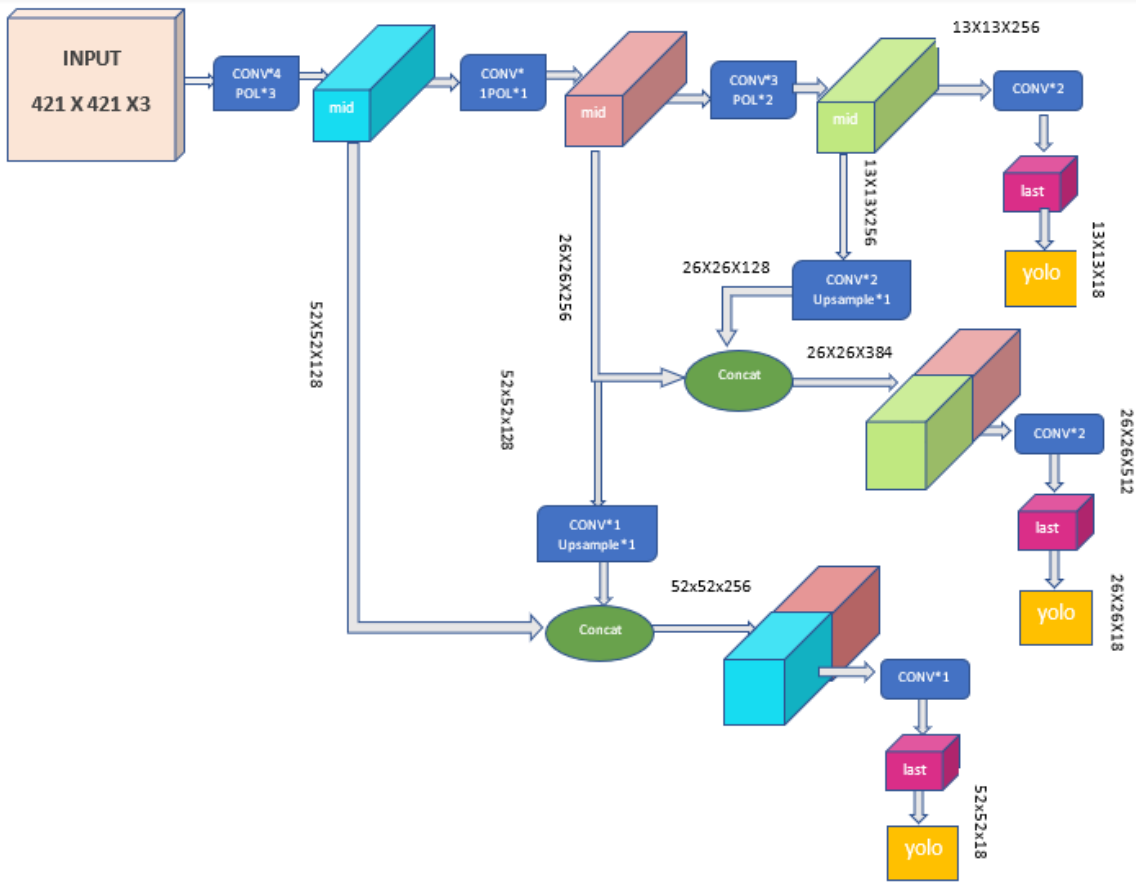

*1) Implementation*

**Built pretrained darknet model for yolov3:** We are using Google Colab so following are the steps we have followed to detect the drones.

First before implementing detection, we will enable the GPU in our notebook so that our YOLOv3 system will be able to process detections over 100 fasters than CPU. We will clone and build darknet by enabling OpenCV and GPU. We then downloaded the pretrained weights of YOLOv3 which is trained on coco dataset having 80 classes that can predict and give detections.

**.Data file and .Name file:** In file drone.data we have provided specifications for our object detector and some significant paths.

In Name file drone.Names, we define the name of the object to be classified and our object to be detected name is drone.

**Configuration Details:** Following are the requirements for training a custom object detector.

YOLOv3 needs a configuration file in which all important training parameters are defined and initialized. In our custom configuration file yolo-drone.cfg, we have declared the parameters.

**Batch Size:** This is a parameter which basically used in the process of training. As we know, the training process is basically a process of updating the weights of the neural network. This is done iteratively and is based on how many faults is our model making on the training set. We cannot use the whole set of images in the training process at once for updating the weights but we use subset (batch size) of the total images used in an iteration. For our custom object detector, we used a batch size of 64. This means that in one iteration, a total of 64 images will be used for updating of the parameters.

**Subdivisions Configuration:** By using subdivision parameter, we use a fraction of batch size at one time if we do not have enough memory to use 64 batch sizes at once. So, we can start training with subdivision =1 and increase its value with the multiple of 2 till the training successfully proceeds. If we specify subdivision parameter then the processing of batch will be performed at any time but processing of all images will only be done when an entire batch or iteration is completed. For our custom object detector, we have used a subdivision of 8 for training set and have used a sub division of 1 for testing set.

**Width, Height and Channels:** These parameters are important as they indicate input image size and the total channels. One of the first steps when it comes to processing the input images which will be used for training is resizing them to the width and height specified. For our custom object detector, we used 416×416 which is the default value. As for the channels we have used 3. Using 3 channels means that there are a total of 3 channel RGB input images being processed.

**Momentum and Decay:** To control how the large weights will be updated between iterations, we use momentum. The weights are being updated after a small batch of images. There is a possibility the weight updates vary quite a bit therefore these parameters are used. Weights can overfit any training data as usually a neural network has very large number of weights. In overfitting, a model will train well on training data and on testing data it does not perform well. It is most likely that the neural network has learnt the answers to all images, but it has not really learned the concept of detecting. To overcome this problem, we penalize large value for weights. The decay also controls this penalty term. For our custom object detector, we have used the value of momentum 0.9 and for decay we have used 0.0005.

**Learning rate:** Learning rate is an important parameter which determines how fast our model will learn. It is defined as the number we multiply by a resulting gradient by. Typically, this is a number between 0.01 and 0.0001. For our custom object detector, we have used a value of 0.001.

The purpose of training is to minimize the loss between outputs i.e. actual and predicted output.

The path to minimize loss is occurring over numerous steps. Training is started with arbitrary set weights and incrementally update the weights until we reach the minimized loss.

**Number of iterations:** In the parameter Max-Batch, we need to specify the number of iterations we should use for- the training process be run for. For the training of our custom object detector, we have used 50,000 number of iterations.

IV. RESULTS

**Evaluation matrix of our model:**
In the tables below, training performance of our custom detector model and previous model is shown:
**Our Custom Architecture Results:**
Batch size 64, subdivisions= 8, Learning rate 0.001 Total test images for validation: 268

| Parameters | Improved Tiny YOLOv3 architecture | Tiny YOLOv3 |
|---|---|---|
| **Average Loss** | 0.22 | 0.28 |
| **Recall** | 0.93 | 0.90 |
| **Precision** | 0.91 | 0.95 |
| **F1-score** | 0.92 | 0.92 |
| **Average IoU** | 70.37% | 73.47% |
| **Average FPS** | 52.6 | 56.6 |
| **mAP@0.5** | 91.8% | 92.1% |
| **TP** | 248 | 240 |
| **FP** | 25 | 13 |
| **FN** | 18 | 26 |

**Table 1 Results Comparison**

```
detections_count = 394, unique_truth_count = 266
class_id = 0, name = Drone, ap = 91.81%        (TP = 248, FP = 25)

for conf_thresh = 0.25, precision = 0.91, recall = 0.93, F1-score = 0.92
for conf_thresh = 0.25, TP = 248, FP = 25, FN = 18, average IoU = 70.37 %

IoU threshold = 50 %, used Area-Under-Curve for each unique Recall
mean average precision (mAP@0.50) = 0.918130, or 91.81 %
Total Detection Time: 2 Seconds
```

Figure 8 Results Evaluated.

```
 FPS:67.3        AVG_FPS:60.6
   cvWriteFrame
 Objects:

 FPS:64.9        AVG_FPS:60.6
   cvWriteFrame
 Objects:

 FPS:66.5        AVG_FPS:60.6
   cvWriteFrame
 Objects:

 FPS:66.4        AVG_FPS:60.6
   cvWriteFrame
 Objects:

 FPS:68.2        AVG_FPS:60.6
 Stream closed.

   cvWriteFrame
 input video stream closed.
   closing... closed!output_video_writer closed.
```

Figure 9 AVG FPS

V. DISCUSSION

We have tested our proposed system on test images and videos and evaluated its performance. To emphasize the performance of our modified lightweight deep learning drone detector-custom tiny yolov3, we have compared its performance with conventional YOLOv3 tiny. The performance of our modified version in terms of precision, recall and frames per second(fps) is improved. The precision, recall and fps of our modified architecture is 0.91, 0.93, 53 which means our architecture is more robust and efficient in its detections.

Average loss means how well an architecture models the data. If the loss is less, the architecture is good at mapping the relationship between given inputs and the target output. Compared to the original Tiny YOLOv3 our Average loss decreased by factor of 0.06.

Recall defines the detector's ability to find all positive instances correctly which means how good detectors will find all the positives. Compared to the original Tiny YOLOv3 our Recall increase by factor of 3% i.e. we achieved recall of 93% while the original architecture has a recall of 90%. .

Our main goal was to improve recall and mAP Mean Average Precision. Because greater recall indicates how detector's ability to find all positive instances correctly which means how good detectors will find all the positives. If we increase the FPS then the detection would be faster as detector would be able to take inputs more fast. The mAP for object detection is the average of the AP calculated for all the classes. mAP@0.5 means that it is the mAP calculated at IOU threshold 0.5.

FPS depends on several factors like input size, computational power, and resolution of video. To increase FPS, the model was trained by reducing the input size and although we achieved greater fps, the recall decreased. There is a general trend that there is a trade-off between Precision and recall thus our precision along with F1-score increased as the recall decreased.

## VI. CONCLUSIONS AND FUTURE WORK

To draw a conclusion, we have made a custom Tiny YOLOv3 architecture which will help in the detection of drones. Our contribution in solving this problem is proposing a solution which is giving a high performance in detecting objects in terms of better precision and recall than its previous versions. We developed this architecture by scaling up the original Tiny YOLOv3 architecture. The original version was performing detection at 2 scale, and we up scaled it by using a route layer which got the features from the previous layers. This architecture now performs detection at 52*52 scale.

Regarding future work, a large dataset containing different types and sizes of drones will be acquired and prepared for real time object detection.

For real time object detection, this task has an extension of hardware implementation which will be carried out in the future. The hardware architecture will be an autonomous system consisting of components including Microcontroller i.e., Raspberry pi and Cameras which will be mounted on the turret interfaced with servo motor. After interfacing the Raspberry pi with all components, the proposed algorithm will be executed. Cameras will be used to get the real time input in the form of a video which will be later converted into frames. These frames will be the input to the main controller for the autonomous system which will detect the object. The rotation of camera will be used for the purpose of tracking. Tracking will be performed with an optimized object tracking algorithm and the camera will rotate in the direction in which the object will move.